\pdfoutput=1
\documentclass[11pt]{article}
\usepackage{acl}
\usepackage{times}
\usepackage{latexsym}
\usepackage[T1]{fontenc}
\usepackage[utf8]{inputenc}
\usepackage{microtype}
\usepackage{inconsolata}
\usepackage{booktabs}
\usepackage{graphicx}
\usepackage{amsmath}

\title{
Knowing the Form, Not the Function:\\
Automatically Auditing Answer--Authority Decoupling in Legal Benchmarks
}

\author{Hsien-Jyh Liao \\
  Ministry of Justice, Taiwan \\
  \texttt{hjliao123@gmail.com}}

\begin{document}
\maketitle

\begin{abstract}
Legal benchmarks typically score final answers even when models also state legal authority. We test whether answer correctness can serve as a proxy for authority grounding. Under ordinary reasoning prompts that did not request statutory citations, four LLMs spontaneously produced authority markers across 238 Taiwan bar-examination items. Because each item has a verified governing provision, we automatically audit answer correctness and authority grounding jointly. The two dimensions dissociate in both directions. In criminal law, 24.0--42.4\% of valid responses were answer-correct but missed the gold authority, while 15.2--21.7\% were answer-incorrect but cited it. A separate statutory-retrieval probe and a permissive citation-abstention intervention further show that answer and citation behavior can move separately at the output level. Because this mismatch arises without adversarial or inconsistency-inducing prompting, answer-only scoring treats naturally occurring gold-authority misses as complete benchmark successes. Because statutory authority is structurally extractable and externally verifiable, the failure can be measured automatically. A preliminary PRC civil-law extension also observes citation-unrequested authority marking, motivating a full cross-jurisdictional joint audit. We therefore propose joint answer--authority evaluation for statute-grounded legal benchmarks.
\end{abstract}

\section{Introduction}
\label{sec:intro}

A model selects the correct answer to a legal benchmark item while citing a provision that does not match the benchmark's verified gold authority. The benchmark records a success. Individual citation errors are unsurprising to legal practitioners; the methodological question is whether they arise systematically within ordinary benchmark responses, whether answer accuracy detects them, and whether they can be measured at scale.

Answer-only scoring implicitly compresses two potentially distinct targets into one. For item $i$, the benchmark usually records only whether the predicted answer $\hat A_i$ matches the gold answer $A_i^*$. Yet a reasoning response may also state a governing authority $\hat G_i$, whose relation to the verified gold authority $G_i^*$ is legally consequential. If answer correctness and gold-authority grounding can come apart, a correct final answer is necessary but not sufficient evidence of a fully successful statute-grounded response.

Taiwan's National Bar Examination provides a compact test of this assumption. Each retained item has a verified answer and governing statutory provision. Models were asked to reason but were never asked to cite a statute; nevertheless, they produced statutory authority markers at high rates. Those markers can be parsed and matched against the gold provision without fresh expert review of every output.

We therefore score the joint state $(A,G)$, where $A$ denotes answer correctness and $G$ denotes gold-authority grounding. Across four models, the two dimensions exhibit a \emph{double dissociation}: correct answers frequently miss the gold authority, and incorrect answers frequently cite it. In criminal law, the two mismatch cells account for 24.0--42.4\% and 15.2--21.7\% of valid responses, respectively. Because the mismatch appears under ordinary, citation-unrequested prompts, it is a validity problem within the benchmark's normal evaluation regime rather than merely an adversarial stress-test failure. Because authority grounding is automatically auditable, the same finding yields a practical correction: report answer accuracy, gold-authority hit rate, and their joint distribution.

Our contributions are threefold. First, we identify a non-induced, automatically auditable double dissociation between answer correctness and authority grounding. Second, we provide a joint scoring protocol that can be added to existing statute-grounded benchmarks. Third, retrieval, non-gold-authority, and citation-abstention analyses show why the two dimensions should be treated separately, without making mechanistic claims about hidden reasoning.

\section{Related Work}
\label{sec:related}

\paragraph{Legal benchmark evaluation.}
Legal NLP benchmarks have progressively broadened what counts as legal capability. LegalBench organizes 162 tasks across multiple forms of legal reasoning, while LawBench evaluates Chinese legal knowledge through 20 tasks at three cognitive levels \cite{guha2023legalbench,fei-etal-2024-lawbench}. More recent work moves beyond coarse task accuracy: PLawBench evaluates realistic legal workflows with fine-grained expert rubrics, and LeMAJ decomposes long legal answers into self-contained Legal Data Points for professional-style automated review \cite{shi2026plawbench,enguehard-etal-2025-lemaj}. These approaches expand the tasks or dimensions being evaluated. They do not, however, test whether the final answer and the legal authority stated in the same ordinary benchmark response are jointly correct.

\paragraph{Reasoning and authority reliability.}
Related work has separately examined legal reasoning quality and citation reliability. Attribution and concept-erasure analyses suggest that accurate legal outcome prediction need not be grounded in legally relevant reasoning \cite{staliunaite-etal-2024-comparative}. Fine-grained legal-text evaluation uses expert taxonomies, trained detectors, or error annotation to identify gaps, hallucinations, citation omissions, and unclear references \cite{hou-etal-2024-gaps,gui-etal-2025-evaluating}. LegalCiteBench directly evaluates citation recovery and verification under citation-focused prompts, while TW-LegalBench reports examination and statute-citation performance across distinct task settings \cite{chen2026legalcitebench,chen2026twlegalbench}. Beyond law, naturally occurring CoT unfaithfulness has been observed without artificially inserted bias \cite{arcuschin-etal-2025-wild}. Our setting combines properties studied separately in these lines: authority markers arise without being requested in an ordinary legal response, have an external gold provision, and can be scored jointly with the answer. The resulting bidirectional mismatch is therefore a directly measurable validity problem for answer-only legal evaluation.

\section{Joint Evaluation Setup}
\label{sec:method}

\subsection{Testbed and Models}
\label{subsec:testbed}
We began with 240 four-option questions from the 2022 and 2025 Taiwan bar examinations: 120 civil-law and 120 criminal-law items. After excluding two civil-law items with source-extraction defects, the retained dataset comprised 238 questions: 118 civil-law and 120 criminal-law items. Each has a verified correct answer and governing statutory provision, derived from two independent commercial solution manuals and checked against the official Laws and Regulations Database of the Republic of China.

We evaluate Gemma-4-31B, LLaMA-3.3-70B, GPT-4o-mini, and Qwen3.6 Flash (\texttt{qwen/qwen3.6-flash}). The models were selected to represent four practically relevant profiles rather than to construct a frontier-performance leaderboard: a locally deployable model (Gemma), a large general-purpose open-weight model (LLaMA), a widely used commercial API baseline (GPT-4o-mini), and a model family selected to provide Chinese-language coverage (Qwen). All experiments use temperature 0.1. C1, C2, and the C2-1 intervention use five random seeds (42--46); C3 and C4 collect one response per item.

\subsection{Measurement Window and Diagnostics}
\label{subsec:protocol}
C2 is the core measurement window: the model receives a legal MCQ and is asked for step-by-step reasoning, but is not asked to cite statutes. The same response yields both answer correctness and authority grounding. C1 is an answer-only baseline; C3 asks the model to identify up to three governing provisions; and C4 probes uncertainty about statutory content. These additional conditions diagnose the observed mismatch but do not define it. Complete prompts appear in Appendix~\ref{app:prompts}.

For each C2 response, $A^+$ denotes a correct final answer and $G^+$ denotes at least one hit on the verified governing authority; $A^-$ and $G^-$ denote the corresponding failures. The evaluation unit is therefore
\[
(A,G)\in\{A^+G^+,A^+G^-,A^-G^+,A^-G^-\}.
\]
The rule-based matcher extracts code name, article number, appended article suffix (e.g., -1), paragraph, and item. Article-level gold accepts a more specific paragraph under the same article; appended articles are matched strictly in both directions, and numeric boundaries prevent matches such as Article 27 with Article 271. A code name may carry over from the preceding local context. C3 is reported primarily as Hit@3 because the prompt permits up to three provisions; first-listed Hit@1 is reported in the appendix.

All response-level metrics use valid model responses. API, transport, and timeout failures that produced no model response are excluded from numerators and denominators and reported separately. Returned outputs remain in the denominator even when answer tags are malformed, citations are unparseable, or no citation is present; they are scored as failures where appropriate. Percentage-point differences are computed from unrounded proportions.

\section{Findings}
\label{sec:findings}

C1 serves only as an answer-accuracy baseline (full values in Appendix~\ref{app:c1}). In civil law, C2 accuracy differs from C1 by only 0.6--3.0 percentage points across models. The criminal-law effect is heterogeneous: accuracy is unchanged for GPT-4o-mini and Qwen, decreases by 4.7 points for Gemma, and decreases by 22.8 points for LLaMA. We therefore do not use the C1--C2 contrast as evidence about the causal role or faithfulness of stated reasoning; the double dissociation below is established entirely within C2.

\subsection{Non-Induced Answer--Authority Double Dissociation}
\label{subsec:dd}
Table~\ref{tab:dd} reports the two mismatch cells. In civil law, 32.7--63.1\% of responses are answer-correct but lack a gold-authority hit; in criminal law the range is 24.0--42.4\%. The reverse mismatch is also present, especially in criminal law: 15.2--21.7\% of responses are answer-incorrect but cite the gold authority. Conditional on a correct criminal-law answer, the authority-miss rate remains 34.9--66.3\% across models.

\begin{table}[t]
\centering
\footnotesize
\setlength{\tabcolsep}{4.5pt}
\begin{tabular}{lrrrr}
\toprule
& \multicolumn{2}{c}{\textbf{Civil law}} & \multicolumn{2}{c}{\textbf{Criminal law}} \\
\cmidrule(lr){2-3}\cmidrule(lr){4-5}
\textbf{Model} & $A^+G^-$ & $A^-G^+$ & $A^+G^-$ & $A^-G^+$ \\
\midrule
Gemma & 63.1 & 2.7 & 42.4 & 15.2 \\
LLaMA & 40.5 & 5.6 & 27.5 & 21.7 \\
GPT-4o-mini & 32.7 & 3.7 & 27.0 & 20.0 \\
Qwen & 33.6 & 12.1 & 24.0 & 21.0 \\
\bottomrule
\end{tabular}
\caption{Answer--authority mismatch cells in C2. $G^-$ includes both absent citations and non-gold citations. Percentages use valid-response denominators.}
\label{tab:dd}
\end{table}

The two directions matter jointly. $A^+G^-$ shows that answer-only scoring records a large class of gold-authority misses as successes. $A^-G^+$ shows that a gold-authority hit is not a substitute for answer correctness. The targets are therefore not merely noisy versions of one another; they must be evaluated jointly.

$G^-$ aggregates two distinct output states: responses that cite no statute and responses that cite only non-gold statutes. Appendix~\ref{app:fullresults} decomposes every joint cell by authority status. The mismatch is not an artifact of silent responses: answer-correct responses whose citations are exclusively non-gold account for 18.0--41.0\% of valid responses across models and domains. Conversely, because $G^+$ requires only one gold hit, it is a recall-style criterion; in every model and domain, gold hits accompanied by additional non-gold citations outnumber clean gold-only responses.

\subsection{Grounding and the Structure of Authority Misses}
\label{subsec:grounding}
Independent statutory retrieval varies sharply across models and domains (Table~\ref{tab:c3main}). Criminal-law Hit@3 exceeds the civil-law rate by 11.7, 24.1, 13.2, and 2.6 percentage points for Gemma, LLaMA, GPT-4o-mini, and Qwen, respectively. Yet the answer--authority mismatch remains substantial in every model.

\begin{table}[t]
\centering
\small
\setlength{\tabcolsep}{5pt}
\begin{tabular}{lrrr}
\toprule
\textbf{Model} & \textbf{Civil Hit@3} & \textbf{Crim. Hit@3} & \textbf{$\Delta$ pp} \\
\midrule
Gemma & 6.8 & 18.5 & +11.7 \\
LLaMA & 5.1 & 29.2 & +24.1 \\
GPT-4o-mini & 6.8 & 20.0 & +13.2 \\
Qwen & 42.4 & 45.0 & +2.6 \\
\bottomrule
\end{tabular}
\caption{C3 statutory retrieval. Hit@3 is the primary measure; differences use underlying unrounded proportions.}
\label{tab:c3main}
\end{table}

Within Qwen's civil-law questions, stronger retrieval is associated with fewer authority misses among correct answers: $P(G^-\mid A^+)$ is 23.3\% on C3-correct questions and 64.1\% on C3-wrong questions, a 40.8-point gap (question-blocked bootstrap 95\% CI [26.0, 55.0]). This is an association, not a causal claim.

Authority misses also retain the appearance of legal support. In Gemma's civil-law outputs, 371 $A^+G^-$ responses contain 548 within-response-deduplicated non-gold citation instances; 241 responses contain at least one citation and 130 contain none. All 548 extracted instances conform to the statutory citation syntax recognized by the parser; 42.2\% fall within 50 articles of the gold provision, and the median numeric distance is 78. Among citation-bearing responses, citation density is nearly the same when grounded and non-gold (2.34 vs.\ 2.27 unique citations per response). Thus, neither citation presence nor citation form reliably signals grounding. A compact five-seed example appears in Appendix~\ref{app:nongold}.

The Admission Paradox remains as a supporting diagnostic: GPT-4o-mini admits uncertainty about every directly probed civil statute in C4, yet produces article citations in 89.7\% of citation-unrequested C2 responses. This reinforces the narrower benchmark claim but is not required to establish the double dissociation.

\subsection{Citation Behavior Can Move without Comparable Answer Change}
\label{subsec:probe}
The C2-1 intervention adds one permissive sentence: when unsure of the exact article number, the model may reason without naming one. It does not prohibit citation. In Gemma, LLaMA, and GPT-4o-mini, article-citation presence falls from 70.7--89.7\% to about 5\%, while answer accuracy changes by at most 0.9 points. Qwen's citation rate also falls substantially, from 96.4\% to 57.3\%, while answer accuracy decreases 2.3 points (Table~\ref{tab:probe_main}).

\begin{table}[t]
\centering
\footnotesize
\setlength{\tabcolsep}{4pt}
\renewcommand{\arraystretch}{1.12}
\begin{tabular}{lrrrr}
\toprule
& \multicolumn{2}{c}{\textbf{Article citation}} & \multicolumn{2}{c}{\textbf{Answer accuracy}} \\
\cmidrule(lr){2-3}\cmidrule(lr){4-5}
\textbf{Model} & \textbf{C2} & \textbf{C2-1} & \textbf{C2} & \textbf{C2-1} \\
\midrule
Gemma & 70.7 & 5.3 & 71.3 & 71.5 \\
LLaMA & 79.5 & 5.1 & 42.5 & 42.5 \\
GPT-4o-mini & 89.7 & 4.9 & 37.1 & 38.0 \\
Qwen & 96.4 & 57.3 & 74.7 & 72.4 \\
\bottomrule
\end{tabular}
\caption{Citation-abstention intervention on civil-law items, five seeds. Citation denotes exact article-pattern presence, not a gold-authority hit.}
\label{tab:probe_main}
\end{table}

The intervention provides output-level evidence that observable answer and citation behavior can be moved separately. It does not identify independent internal mechanisms or establish that every suppressed citation was ungrounded.

\section{Discussion}
\label{sec:discussion}
The practical phenomenon---a correct legal answer accompanied by a non-gold authority---is familiar. The contribution is to formalize it as a benchmark-validity problem. Under ordinary, citation-unrequested conditions, the mismatch is systematic, bidirectional, and measurable without case-by-case expert scoring. Answer accuracy therefore remains necessary but is not sufficient for statute-grounded legal evaluation.

A minimal benchmark correction follows directly. Where a task has verified governing authorities, evaluation should report answer accuracy, authority Hit@$k$, and the joint distribution $A^+G^+$/$A^+G^-$/$A^-G^+$/$A^-G^-$. Citation absence should also be separated from citation-bearing non-gold authority. Where responses contain multiple citations, benchmarks should further distinguish clean gold-only support from mixed gold-plus-non-gold support, or report citation-level precision alongside gold-authority recall. This is not a proposal to replace answer accuracy or to score the full legal validity of every reasoning step. It is a low-cost extension that prevents answer-only scoring from treating every correct option as a complete success.

The broader implication concerns stated reasoning rather than hidden causal computation. Statutory authority is one externally verifiable component of legal justification. Its dissociation from final answers shows that fluent legal form cannot be treated as evidence that the asserted authority supports the conclusion.

Qwen's higher statutory-retrieval performance and lower gold-authority miss rates show that the magnitude of the mismatch varies substantially across models and may be mitigated by differences in model training or legal-domain coverage. This heterogeneity complements rather than undermines joint scoring: the proposed audit is what makes it possible to measure whether such improvements actually translate into better answer--authority alignment.

A preliminary extension further finds citation-unrequested authority marking in PRC civil-law responses (Appendix~\ref{app:china}), motivating a future joint answer--authority audit beyond the Taiwan-law testbed.

\section{Conclusion}
\label{sec:conclusion}
We identify a non-induced, automatically auditable double dissociation between answer correctness and authority grounding in ordinary legal benchmark responses. The finding turns a familiar practical concern into a benchmark-level methodological claim: answer-only evaluation systematically hides gold-authority misses, yet statute-grounded tasks already contain the structure needed to measure them. Legal benchmarks should therefore score the answer and its asserted governing authority jointly.

\section*{Limitations}
Our claims are behavioral, not mechanistic. Authority matching evaluates a verifiable component of stated legal reasoning, not the full validity or causal role of the reasoning chain. Strict gold matching may count legally relevant alternative provisions as misses, although gold provisions were independently verified. The approach is most directly applicable to codified, citation-normed tasks; common-law and open-ended settings may require authority sets or expert annotation. Because multiple-choice questions permit option elimination and other answer-selection paths that do not require explicit recovery of the governing rule, the magnitude of the observed mismatch may be larger than in open-ended legal tasks; direct replication in open-ended settings is needed. The model set was designed to cover practically relevant deployment profiles---local deployment, open-weight general-purpose use, a commercial API baseline, and a Chinese-oriented model family---rather than to exhaustively represent the latest frontier or deliberative reasoning models; whether such models exhibit lower rates of answer--authority mismatch remains an open empirical question. The Mainland Chinese extension establishes only the presence of spontaneous authority marking; it does not yet replicate the joint answer--authority analysis.

\section*{Ethical Considerations}
The study uses public professional-examination questions and contains no personal or confidential case data. It evaluates model outputs rather than deploying an automated legal decision system. The proposed authority audit is an evaluation tool, not a substitute for professional review of legal advice. Commercial solution manuals were used only to construct and verify the gold set; their text is not redistributed with this submission.

\appendix

\section{Testbed Construction}
\label{app:testbed}

\subsection*{Question Format and Gold Construction}
Each question presents a legal scenario followed by four answer options. Correct answers and governing statutes were derived from two independent commercial solution manuals and cross-checked against the official Laws and Regulations Database of the Republic of China. Each retained gold provision was confirmed to exist, to address the legal issue in the item, and to be the provision most directly applicable to the correct option. The initial collection contained 240 items. Two civil-law items had source-extraction defects and were excluded before analysis, leaving 118 civil-law and 120 criminal-law items. No retained gold answer or statute was contradicted by either manual.

\begin{table}[h]
\centering
\scriptsize
\setlength{\tabcolsep}{2pt}
\resizebox{\columnwidth}{!}{%
\begin{tabular}{lrrrl}
\toprule
\textbf{Domain} & \textbf{Year} & \textbf{Raw N} & \textbf{Retained N} & \textbf{Statutes covered} \\
\midrule
Civil & 2022 & 60 & 59 & Civil Code; Civil Procedure Code \\
Civil & 2025 & 60 & 59 & Civil Code; Civil Procedure Code \\
Criminal & 2022 & 60 & 60 & Criminal Code; Criminal Procedure Code \\
Criminal & 2025 & 60 & 60 & Criminal Code; Criminal Procedure Code \\
\midrule
Total & & 240 & 238 & \\
\bottomrule
\end{tabular}%
}
\caption{Dataset composition before and after the two documented source-extraction exclusions.}
\label{tab:dataset}
\end{table}

\subsection*{Human Citation Norm}
The commercial solutions conventionally cite statutory provisions when explaining an answer. This motivates using C2 as a citation-unrequested but citation-normed setting: the prompt requests reasoning, not an article number, while the professional register makes spontaneous authority marking observable.

\subsection*{Valid-Response Denominators}
Civil-law C2 has 2,357 valid responses: Gemma 588, LLaMA 590, GPT-4o-mini 590, and Qwen 589. Criminal-law C2 has 2,394 valid responses: Gemma 594 and 600 for each other model. No-response API, transport, or timeout failures are excluded; returned format, parse, and zero-citation outputs remain.

\section{Complete Joint and Diagnostic Results}
\label{app:fullresults}

\subsection*{Civil-Law Joint Distribution}
\begin{table}[h]
\centering
\small
\resizebox{\columnwidth}{!}{%
\begin{tabular}{lrrrrr}
\toprule
\textbf{Model} & $A^+G^+$ & $A^+G^-$ & $A^-G^+$ & $A^-G^-$ & $P(G^-\mid A^+)$ \\
\midrule
Gemma & 8.2 & 63.1 & 2.7 & 26.0 & 88.5 \\
LLaMA & 2.0 & 40.5 & 5.6 & 51.9 & 95.2 \\
GPT-4o-mini & 4.4 & 32.7 & 3.7 & 59.2 & 88.1 \\
Qwen & 41.1 & 33.6 & 12.1 & 13.2 & 45.0 \\
\bottomrule
\end{tabular}%
}
\caption{Civil-law C2 joint distribution. Minor row-sum variation may arise from displayed one-decimal rounding.}
\label{tab:civiljoint}
\end{table}

\subsection*{Criminal-Law Joint Distribution}
\begin{table}[h]
\centering
\small
\resizebox{\columnwidth}{!}{%
\begin{tabular}{lrrrrr}
\toprule
\textbf{Model} & $A^+G^+$ & $A^+G^-$ & $A^-G^+$ & $A^-G^-$ & $P(G^-\mid A^+)$ \\
\midrule
Gemma & 21.6 & 42.4 & 15.2 & 20.8 & 66.3 \\
LLaMA & 18.5 & 27.5 & 21.7 & 32.3 & 59.8 \\
GPT-4o-mini & 17.5 & 27.0 & 20.0 & 35.5 & 60.7 \\
Qwen & 44.8 & 24.0 & 21.0 & 10.2 & 34.9 \\
\bottomrule
\end{tabular}%
}
\caption{Criminal-law C2 joint distribution. Minor row-sum variation may arise from displayed one-decimal rounding.}
\label{tab:criminaljoint}
\end{table}

\subsection*{Authority-Status Decomposition of the Joint Cells}
Each response is classified by authority status: no citation, only non-gold citations, only gold-matching citations, or gold plus non-gold citations. Table~\ref{tab:fourway} crosses this decomposition with answer correctness.

\begin{table}[h]
\centering
\scriptsize
\setlength{\tabcolsep}{2.5pt}
\resizebox{\columnwidth}{!}{%
\begin{tabular}{llrrrrrrrr}
\toprule
& & \multicolumn{4}{c}{\textbf{Civil}} & \multicolumn{4}{c}{\textbf{Criminal}} \\
\cmidrule(lr){3-6}\cmidrule(lr){7-10}
\textbf{Model} & & None & NG & G & G{+}NG & None & NG & G & G{+}NG \\
\midrule
Gemma & $A^+$ & 22.1 & 41.0 & 3.2 & 4.9 & 15.3 & 27.1 & 6.4 & 15.2 \\
      & $A^-$ & 3.9 & 22.1 & 1.5 & 1.2 & 5.1 & 15.8 & 3.0 & 12.1 \\
LLaMA & $A^+$ & 12.5 & 28.0 & 0.8 & 1.2 & 9.5 & 18.0 & 10.2 & 8.3 \\
      & $A^-$ & 8.0 & 43.9 & 0.8 & 4.7 & 13.3 & 19.0 & 7.7 & 14.0 \\
GPT-4o-mini & $A^+$ & 3.2 & 29.5 & 1.5 & 2.9 & 2.5 & 24.5 & 10.3 & 7.2 \\
      & $A^-$ & 7.1 & 52.0 & 0.3 & 3.4 & 4.5 & 31.0 & 6.8 & 13.2 \\
Qwen & $A^+$ & 1.5 & 32.1 & 3.7 & 37.4 & 2.2 & 21.8 & 6.0 & 38.8 \\
      & $A^-$ & 1.0 & 12.2 & 1.5 & 10.5 & 0.5 & 9.7 & 2.2 & 18.8 \\
\bottomrule
\end{tabular}%
}
\caption{Authority-status decomposition of C2 responses (\% of valid responses). None: no citation; NG: non-gold citations only; G: gold-matching citations only; G{+}NG: gold plus non-gold citations. A response is citation-bearing when the matcher extracts at least one parseable statutory citation; this criterion is marginally more permissive than the surface article-pattern rate in Table~\ref{tab:probe_main} (differences $\leq$3.3 points).}
\label{tab:fourway}
\end{table}

\subsection*{C3 Statutory Retrieval and C4 Civil-Law Admission}
\begin{table}[h]
\centering
\footnotesize
\setlength{\tabcolsep}{4pt}
\resizebox{\columnwidth}{!}{%
\begin{tabular}{llrrr}
\toprule
\textbf{Model} & \textbf{Domain} & \textbf{C3 Hit@3} & \textbf{C3 first Hit@1} & \textbf{C4 Admit} \\
\midrule
Gemma & Civil & 6.8 & 6.8 & 9.3 \\
LLaMA & Civil & 5.1 & 4.2 & 92.4 \\
GPT-4o-mini & Civil & 6.8 & 5.9 & 100.0 \\
Qwen & Civil & 42.4 & 33.9 & 10.2 \\
\midrule
Gemma & Criminal & 18.5 & 16.8 & --- \\
LLaMA & Criminal & 29.2 & 19.2 & --- \\
GPT-4o-mini & Criminal & 20.0 & 15.8 & --- \\
Qwen & Criminal & 45.0 & 33.3 & --- \\
\bottomrule
\end{tabular}%
}
\caption{C3 retrieval (both domains) and C4 admission (civil law only). C3 first Hit@1 denotes the first extracted citation, not a learned ranking metric. C4 retains the original probe definition; only civil-law C4 is reported here because its valid-response denominators are fully audited.}
\label{tab:diagnostics}
\end{table}

Civil C3 hits are 8/118, 6/118, 8/118, and 50/118. Criminal hits are 22/119, 35/120, 24/120, and 54/120; one Gemma criminal API failure is excluded. The corresponding cross-domain changes are +11.7, +24.1, +13.2, and +2.6 points.

\subsection*{Qwen Grounding-Stratified Analysis}
Among Qwen's 118 civil-law questions, 50 are C3-correct and 68 are C3-wrong. The authority-miss rate among answer-correct C2 responses is 23.3\% in the former group and 64.1\% in the latter, a 40.8-point gap. A question-level blocked bootstrap with 5,000 resamples gives a 95\% confidence interval of [26.0, 55.0]. The result is consistent with stronger statutory grounding being associated with fewer authority misses; it does not establish causation.

A separate coupling statistic, $\Delta=P(G^+\mid A^+)-P(G^+\mid A^-)$ computed within each group, is $-9.3$ points for C3-correct questions and $+3.8$ points for C3-wrong questions. The between-group difference is $-13.2$ points, with question-level blocked-bootstrap 95\% CI $[-33.2, 9.0]$; no statistically significant difference is detected for that statistic.

\section{The Structure of Non-Gold Authority}
\label{app:nongold}

\subsection*{Aggregate Structure}
Gemma's civil-law C2 outputs contain 371 $A^+G^-$ responses. Of these, 241 contain at least one statutory citation and 130 contain none. Across the 371 responses, the matcher extracts 548 within-response-deduplicated non-gold citation instances. All conform to the statutory citation syntax recognized by the parser; 42.2\% are within 50 article numbers of the gold provision, and the median numeric distance is 78. Among citation-bearing responses, grounded responses contain 2.34 within-response-unique citations on average and non-gold responses contain 2.27. Across all 371 non-gold responses, including the 130 zero-citation cases, the average is 1.48. Comparisons of citation density use the citation-bearing denominator (2.34 vs.\ 2.27); 1.48 is reported only as the overall density including zero-citation responses.

\subsection*{Q42 Five-Seed Example}
For TW\_Bar\_111\_Civil\_Q42, the gold authority is Civil Code Article 1146. Gemma selects the correct answer in all five seeds, never cites Article 1146, and produces 15 non-gold statutory citations across the five outputs. The phrase ``15 citations'' does not imply 15 globally distinct statutes after cross-seed deduplication. Across the ten seed pairs, the mean pairwise Jaccard similarity between cited-statute sets is 0.09 (median 0; maximum 0.50; seven of ten pairs share no statute). The failure is thus systematic at the level of missing the gold authority but stochastic in the particular non-gold authorities used to realize it.

\begin{table}[h]
\centering
\small
\begin{tabular}{clc}
\toprule
\textbf{Seed} & \textbf{Statutes cited in reasoning} & \textbf{Answer} \\
\midrule
42 & \S1101, \S1122, \S1123 & correct \\
43 & \S1103, \S1129, \S1130 & correct \\
44 & \S1101-1, \S1129, \S1212 & correct \\
45 & \S1101, \S1109, \S1123 & correct \\
46 & \S1129-1, \S101, \S1122 & correct \\
\midrule
\multicolumn{3}{l}{\textit{Gold: \S1146, absent from every seed}}\\
\bottomrule
\end{tabular}
\caption{A repeated $A^+G^-$ case. The example illustrates unstable non-gold authority, not five independent questions.}
\label{tab:q42}
\end{table}

\section{Prompt Templates}
\label{app:prompts}
The prompts below are faithful English translations of the original Traditional Chinese prompts. For criminal-law evaluations, references to the Civil Code and Code of Civil Procedure were replaced with the Criminal Code and Code of Criminal Procedure.

\subsection*{C1: MCQ Only (Task Accuracy Baseline)}
\begin{quote}
\small
You are a Taiwanese legal expert. Answer the following question based on Taiwan's Civil Code and Code of Civil Procedure. Do not provide any reasoning or explanation. Output only \texttt{<ANSWER>X</ANSWER>}, where X is A, B, C, or D.

[Case Description]: \textit{[question text]}

[Options]: \textit{[(A)... (D)...]}
\end{quote}
C1 is an ablation baseline to C2 and explicitly forbids discourse generation.

\subsection*{C2: MCQ with Reasoning Required}
\begin{quote}
\small
You are a Taiwanese legal expert. Analyze the following case based on Taiwan's Civil Code and Code of Civil Procedure. You must provide step-by-step legal reasoning in Traditional Chinese before answering. No external references are provided; use your own legal knowledge. End exactly with \texttt{<ANSWER>X</ANSWER>}, where X is A, B, C, or D.

[Case Description]: \textit{[question text]}

[Options]: \textit{[(A)... (D)...]}
\end{quote}
No statutory citation is requested. Citations appearing in model outputs are therefore citation-unrequested authority markers.

\subsection*{C3: Forward Grounding (Fact $\to$ Statute ID)}
\begin{quote}
\small
The following is a factual description of a legal case, with the answer options removed: \textit{[fact-only description]}. Which specific article of Taiwan law applies to the core issue? List the statute identifier directly, provide at most three provisions in order of importance, and give no explanation. If you truly do not know, answer ``Uncertain.''
\end{quote}

\subsection*{C4: Epistemic Commitment (Statute ID $\to$ Content)}
\begin{quote}
\small
Recite the complete content of the following Taiwan statute: \textit{[statute identifier]}. List every paragraph and item. Mark any wording about which you are uncertain. If you do not remember the provision, answer that you are uncertain about its exact content. Do not guess or fabricate statutory text.
\end{quote}

\subsection*{C4 Classification Rubric}
C4 responses were classified automatically using strict Chinese keyword rules. \textsc{admits-uncertainty} covers unambiguous statements equivalent to ``I am not sure,'' ``I do not know,'' ``I cannot confirm the exact wording,'' or ``I forgot.'' Legal terms that merely contain the lexical form for uncertainty were excluded. All other responses that supplied statutory content were classified as \textsc{confident-output}, regardless of accuracy. Two human coders validated a random sample of 50 responses (Cohen's $\kappa=0.94$).

\section{Citation-Abstention Probe: Complete Results}
\label{app:probe}
C2-1 adds a permissive sentence allowing reasoning without an exact article number when the model is unsure. All other settings match C2.

\begin{table}[h]
\centering
\footnotesize
\setlength{\tabcolsep}{4pt}
\renewcommand{\arraystretch}{1.12}
\begin{tabular}{lrrrr}
\toprule
& \multicolumn{2}{c}{\textbf{Article citation rate}} & \multicolumn{2}{c}{\textbf{Answer accuracy}} \\
\cmidrule(lr){2-3}\cmidrule(lr){4-5}
\textbf{Model} & \textbf{C2} & \textbf{C2-1} & \textbf{C2} & \textbf{C2-1} \\
\midrule
Gemma & 70.7 & 5.3 & 71.3 & 71.5 \\
LLaMA & 79.5 & 5.1 & 42.5 & 42.5 \\
GPT-4o-mini & 89.7 & 4.9 & 37.1 & 38.0 \\
Qwen & 96.4 & 57.3 & 74.7 & 72.4 \\
\bottomrule
\end{tabular}
\caption{Complete citation-abstention results. Qwen's C2-1 citation rate is 50.1\% on the 2022 items and 64.4\% on the 2025 items.}
\label{tab:probe}
\end{table}

Qwen is a boundary case rather than a contradiction: its civil C3 Hit@3 is 42.4\%, substantially above the 5.1--6.8\% range of the other three models, and it retains more citation behavior when abstention becomes permissible. This comparison is descriptive; it does not prove that retained citations are grounded. Among the three models whose exact article naming falls to about 5\%, 85.7\% of post-collapse responses still invoke legal authority in non-citational form (for example, ``under the Civil Code'' or ``under the relevant provisions''). The intervention therefore changes the observable, verifiable marker more than it changes answer selection. We interpret this as output-level behavioral separability, not as evidence of independent hidden mechanisms.

\section{Answer-Only Baseline (C1)}
\label{app:c1}
C1 requests only the final answer, with no reasoning. It functions solely as a task-accuracy baseline. Table~\ref{tab:c1c2} reports answer accuracy under C1 and the reasoning-required C2 condition. In civil law the two conditions are close; in criminal law the effect is heterogeneous, most notably a 22.8-point drop for LLaMA. The contrast is reported for completeness and is not used to infer the causal role or faithfulness of stated reasoning.

\begin{table}[h]
\centering
\small
\begin{tabular}{lrrrr}
\toprule
& \multicolumn{2}{c}{\textbf{Civil}} & \multicolumn{2}{c}{\textbf{Criminal}} \\
\cmidrule(lr){2-3}\cmidrule(lr){4-5}
\textbf{Model} & \textbf{C1} & \textbf{C2} & \textbf{C1} & \textbf{C2} \\
\midrule
Gemma & 69.5 & 71.3 & 68.7 & 64.0 \\
LLaMA & 41.9 & 42.5 & 68.8 & 46.0 \\
GPT-4o-mini & 35.3 & 37.1 & 44.5 & 44.5 \\
Qwen & 71.7 & 74.7 & 68.8 & 68.8 \\
\bottomrule
\end{tabular}
\caption{Answer accuracy under the answer-only C1 and reasoning-required C2 conditions.}
\label{tab:c1c2}
\end{table}

\section{Preliminary Observation in Mainland Chinese Civil Law}
\label{app:china}
To examine whether citation-unrequested authority marking is specific to the Taiwan-law testbed, we conducted a preliminary extension using 30 Mainland Chinese civil-law examination questions based on the PRC Civil Code. The same ordinary reasoning prompt was applied to four models, without requesting statutory citations.

Statutory authority markers appeared in 89 of 90 analyzed responses for each of Gemma, LLaMA, and GPT-4o-mini (98.9\%), and in all 88 available Qwen responses (100.0\%). These results show that spontaneous authority marking is also observable in a second Chinese-language statutory jurisdiction.

This extension does not establish a replication of Answer--Authority Decoupling. The present analysis does not yet apply the Taiwan matcher and joint four-cell scoring protocol to the PRC data, and two Qwen outputs were unavailable. We therefore report only citation presence and make no claim about cross-jurisdictional effect sizes or corpus-density mechanisms. A complete answer--authority audit of Mainland Chinese legal benchmarks is left to future work.

\end{document}